\documentclass[10pt,twocolumn,letterpaper]{article}

\pdfoutput=1

\usepackage{iccv}
\usepackage{times}
\usepackage{epsfig}
\usepackage{graphicx}
\usepackage{amsmath}
\usepackage{amssymb}
\usepackage{algorithm}
\usepackage{algpseudocode}

\usepackage[colorinlistoftodos,prependcaption]{todonotes}

\usepackage[pagebackref=true,breaklinks=true,letterpaper=true,colorlinks,bookmarks=false]{hyperref}

\iccvfinalcopy % *** Uncomment this line for the final submission

\def\iccvPaperID{7} % *** Enter the ICCV Paper ID here

\ificcvfinal\fi

\usepackage{xcolor}

\newcommand\blfootnote[1]{%
  \begingroup
  \renewcommand\thefootnote{}\footnote{#1}%
  \addtocounter{footnote}{-1}%
  \endgroup
}

\begin{document}

%%%%%%%%% TITLE
\title{SSIG: A Visually-Guided Graph Edit Distance for Floor Plan Similarity}

\author{
Casper C. J. van Engelenburg\textsuperscript{1, 2}, \quad
Seyran Khademi\textsuperscript{1, 2}, \quad
Jan C. van Gemert\textsuperscript{2} \\
\textsuperscript{1}AiDAPT Lab, 
\textsuperscript{2}Computer Vision Lab, 
Delft University of Technology \\
{\small \texttt{\{c.c.j.vanengelenburg, s.khademi, j.c.vangemert\}@tudelft.nl}}
}

% \author{
% Casper C. J. van Engelenburg\\
% Delft University of Technology
% \and
% Seyran Khademi\\
% Delft University of Technology
% \and
% Jan C. van Gemert \\
% Delft University of Technology \\
% (c.c.j.vanengelenburg; s.khademi, j.c.vangemert)@tudelft.nl
% }

\maketitle
\ificcvfinal\thispagestyle{empty}\fi

% \tableofcontents

%%%%%%%%%% add sections -> to reduce clutter: include per section when writing the piece.
% !!!!!! MAKE SURE TO ADD ALL THE SECTION'S TEXT BELOW (INSTEAD OF \include) FOR THE FINAL SUBMISSION !!!!!
\begin{abstract}

\blfootnote{This paper is published as a workshop paper at ICCV 2023}

We propose a simple yet effective metric that measures structural similarity between visual instances of architectural floor plans, without the need for learning. Qualitatively, our experiments show that the retrieval results are similar to deeply learned methods. Effectively comparing instances of floor plan data is paramount to the success of machine understanding of floor plan data, including the assessment of floor plan generative models and floor plan recommendation systems. Comparing visual floor plan images goes beyond a sole pixel-wise visual examination and is crucially about similarities and differences in the shapes and relations between subdivisions that compose the layout. Currently, deep metric learning approaches are used to learn a pair-wise vector representation space that closely mimics the structural similarity, in which the models are trained on similarity labels that are obtained by Intersection-over-Union (IoU). To compensate for the lack of structural awareness in IoU, graph-based approaches such as Graph Matching Networks (GMNs) are used, which require pairwise inference for comparing data instances, making GMNs less practical for retrieval applications. In this paper, an effective evaluation metric for judging the structural similarity of floor plans, coined \textbf{SSIG} (\textbf{S}tructural \textbf{S}imilarity by \textbf{I}oU and \textbf{G}ED), is proposed based on both image and graph distances. In addition, an efficient algorithm is developed that uses SSIG to rank a large-scale floor plan database. Code will be openly available.
\end{abstract}
\section{Introduction}

% Introduction to floor plans and why they are important in architectural design
Floor plans, as one of the most celebrated media for communicating and thinking in architectural design, are horizontal orthographic projections of a particular building's floor. Floor plans are simple but powerful representations of space, conveying a richness of information about the compositional structure of buildings -- the shapes, proportions, and relations between the building's individual subdivisions. The compositional structure of a building is a key factor in the architectural quality.

% Introduction to data-driven research on floor plans 
The multi-modal nature of floor plans as pictorial (as images), graphical (as a set of geometrical entities), and topological (as graphs) entities, leverage different data modeling and machine learning frameworks for digital representation and machine understanding in the domain of architectural design. In particular, floor plan synthesis, such as floor plan generation \cite{nauata_house-gan_2020, nauata_house-gan_2021, shabani_housediffusion_2022} or style transfer \cite{chaillou_ai_2019}, has often been explored. Another major research line is that of architectural element recognition in which the goal is to automatically learn to extract geometrical features from floor plan images \eg walls or furniture \cite{liu_raster--vector_2017, hampali_monte_2021, rahbar_architectural_2022}.

% Introducing the need for comparing floor plans digitally, and why it is so difficult
Despite the developments in the generative models for floor plans, there have been few valuable attempts at developing data-driven methods to \textit{reason} about floor plans, such as floor plan retrieval by effectively comparing data instances \ie measuring similarity.  As found by \cite{patil_layoutgmn_2021}, structural similarity judgment for floor plans is a challenging task -- it is multi-faceted and is about the shapes and proportions, visual features, and relations between the floor plan's subdivisions. Generally speaking, comparing floor plans is about finding commonalities in the underlying structure, hence referred to as \textit{structural similarity}.

% State-of-the-art approaches to compute structural similarity
The most effective methods for measuring structural similarity between floor plans rely on deeply learned vector representations with the goal to make the feature vector space reflect structural similarity \ie the distance between feature vectors is small for structurally similar floor plans and vice versa. To train such models, \cite{vedaldi_learning_2020} propose to use weakly supervised deep metric learning setup by obtaining similarity labels through Intersection-over-Union (IoU). To compensate for the lack of structural awareness in IoU, \cite{patil_layoutgmn_2021, jin_shrag_2022} use Graph Matching Networks (GMNs) \cite{li_graph_2019} that naturally embed topological information in the feature vectors. Besides the fact that GMNs hinder the practical usability \cite{shih_floor_2022}, we believe that simpler frameworks could learn similarly effective vector representations when the objective is better aligned with structural similarity in the first place. In our work, we develop an improved objective and opt for simplicity. We start by investigating the merits and bottlenecks of well-known image- and graph-based similarity metrics.

% IoU and GED and their relation to structural similarity cognitively and why indepedently they lack to robustly capture structural similarity
The \textit{Intersection-over-Union} (IoU) is often used as an evaluation metric to measure structural similarity \eg to find similar or matching pairs for deep metric learning models \cite{patil_layoutgmn_2021, jin_shrag_2022}, to evaluate the accuracy of structural reasoning algorithms \cite{liu_raster--vector_2017}, or to evaluate the output diversity of generative models \cite{nauata_house-gan_2020}. In this paper, we show how IoU fails to robustly measure structural similarity because 1) IoU is overly sensitive to geometric perturbations in the image representation of a floorplan and 2) IoU is often unable to identify important differences in the connectivity between space subdivisions such as doors, walls, openings. 

Floor plans have been successfully modelled as graphs \cite{nauata_house-gan_2020,patil_layoutgmn_2021}, which allows for graph matching algorithms to compute a distance (similarity) between a pair of graphs. For example, the \textit{Graph Edit Distance} (GED) computes the minimum cost of converting a source graph into another (isomorphic to it) target graph \cite{sanfeliu_distance_1983}. For example, \cite{nauata_house-gan_2020} uses GED to check the compatibility between generated layouts. Although GED effectively captures topological similarity and with that addresses some of IoU's limitations, we show that GED has inherent limitations when used for measuring structural similarity. 

% IoU and GED are not good enough. We introduce SIGG a combination of the two.
Independently, image- and graph-based similarity metrics lack to holistically capture the structural similarity between floor plans. Therefore, instead of treating images and graphs in isolation, we propose an evaluation metric that ascertains closeness based on the floor plans' corresponding image \textit{and} graph, called \textbf{SSIG}. In addition, an efficient algorithm based on \textbf{SSIG} is used to rank RPLAN \cite{wu_data-driven_2019}, a large-scale floor plan database, on structural similarity. Our contributions include:

% Contributions
\begin{itemize}
    \item A study on the distributions and correlations between IoU and GED on a large-scale floor plan database, with one main finding: IoU and GED independently fail to robustly capture structural similarity.
    \item A proposed simple yet powerful measure for the structural similarity of floor plans.
    \item An effective ranking technique to sort a large floor plan database, hence developing a starting point for proper evaluation and training of floor plan retrieval models.
\end{itemize}
\section{Related works}

% Floor plan representations
\paragraph{Floor plan representations.}
Floor plans are digitally represented in various ways that each emphasize different components of interest \eg as images allowing to model fine-grained details such as materialization and furniture \cite{sharma_daniel_2017, khade_rotation_2021}, geometries that explicitly model the shapes, proportions and locations of the shapes of the subdivisions \cite{patil_layoutgmn_2021}, and graphs that model the relations between the subdivisions \cite{nauata_house-gan_2020, patil_layoutgmn_2021, shabani_housediffusion_2022, park_floor_2023}.

\paragraph{Floor plan analysis.}
Traditional approaches on floor plan analysis -- or similar data such as documents \cite{zhong_publaynet_2019} or UIs \cite{deka_rico_2017} -- involve primitive heuristics to approximate a floor plan distance and are, hence, instance-specific \cite{ogorman_executive_1997}. More recently, two-step approaches became the standard to approximate similarity of floor plans \cite{yamada_graph_2021, sabri_semantic_2017, sharma_unified_2016, takada_similar_2018, sharma_daniel_2017, sharma_high-level_2018, weber_scatch_2010, park_floor_2023}, in which the first step involves extracting relevant features from an image \eg a graph representation of the room connectivity \cite{park_floor_2023} and the second step computes a distance based on the extracted features \eg a subgraph matching algorithm for graph similarity \cite{sabri_semantic_2017}. Besides the fact that the two-step approaches are prone to error propagation, the two-step approaches assume that the extracted features, such as the access graph, completely and accurately capture all aspects of floor plans. In our work, we challenge this assumption, particularly concerning semantic image and access graph representations of floor plans.

\paragraph{Structural similarity by Graph Neural Networks.}
Recently, Graph Neural Networks (GNN) have been proposed to \textit{learn} floor plan similarity \cite{patil_layoutgmn_2021, vedaldi_learning_2020, jin_shrag_2022}, learning floor plan vector representations in an end-to-end fashion. Most successfully, \cite{jin_shrag_2022, patil_layoutgmn_2021} leverage Graph Matching Networks (GMN) \cite{li_graph_2019} in combination with weakly-supervised labels based on the IoU. Due to the cross-graph information sharing in GMNs, \cite{patil_layoutgmn_2021} found that GMNs compensate for the lack of structural awareness in IoU and naturally learn to embed the structural commonalities in the vector representations. A critical shortcoming of GMNs is that vector representations cannot be computed in isolation, therefore limiting the practical usability. We believe that the main reason that cross-graph information sharing is needed is that essentially the wrong objective is injected -- that is to mimic a distance metric inspired by IoU. We show that our metric, which is a combination between visual- and a graph-inspired similarity scores, can be leveraged to rank a floor plan database for which the retrievals share similar characteristics as to the works in \cite{patil_layoutgmn_2021}.

\section{Method}

In this section, we develop a simple yet effective strategy to numerically approximate structural similarity between floor plans. Formally, we seek a \textit{similarity function} $s \left(\cdot, \cdot \right)$ between two floor plans $p_1$ and $p_2$:

\begin{equation}\label{eq:sim}
   s \left(p_1, p_2 \right): \mathcal{P} \times \mathcal{P} \rightarrow \mathbb{R}^+,
\end{equation}

\noindent in which $\mathcal{P}$ is the (mathematical) space of a floor plan. $\mathcal{P}$ can be multi-modal \eg a joint space of images and graphs: $\mathcal{P} = \mathbb{R}^{H \times W \times 3} \times \mathcal{G}$. The goal is to make $s$ closely mimic the structural similarity between floor plans: the larger the structural similarity the larger $s$ and vice versa.

The section is split into three main parts. The first part explores the modeling and attribution of floor plans, (Subsec. \ref{sec:3_representation}). The second part investigates known and frequently used image- and graph-based similarity metrics (Subsec. \ref{sec:3_iou}, \ref{sec:3_ged}). The third part describes the new similarity metric (Subsec. \ref{sec:3_ssig}), and provides an algorithm that can efficiently rank a floor plan dataset (Subsec. \ref{sec:3_algo}).

%% FLOOR PLAN DATA AND ATTRIBUTES
\subsection{Floor plan representations}\label{sec:3_representation}

In this work, we consider three well-known floor plan representations: 1) the pictorial image, 2) the semantic image, and 3) the access graph.

\begin{figure}[ht]
    \centering
    \includegraphics[width=0.95\linewidth]{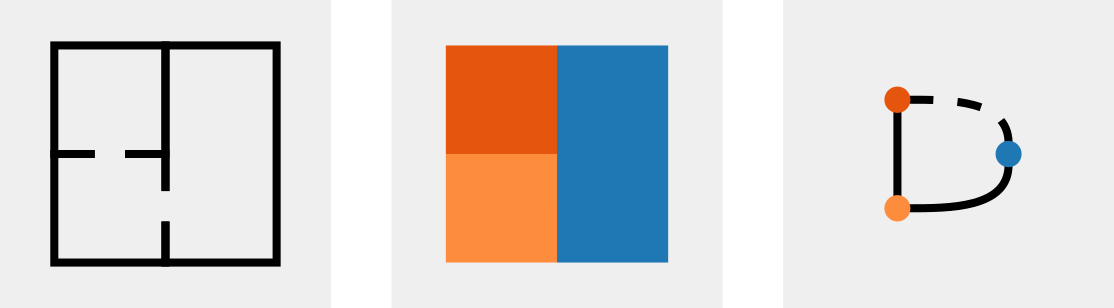}
    \caption{
    \textbf{Floor plan representations}. Left: Pictorial image. Center: Semantic image. Right: Access graph.
    }
    \label{fig:fp-representation}
\end{figure}

\paragraph{Pictorial image} Conventionally, floor plans are modeled as images. All floor plan images that are either grayscale or RGB and do not contain any categorial information at pixel level about the semantics of the subdivisions are referred to as \textit{pictorial images}, $I \in \mathcal{I} = \mathbb{R}^{H \times W (\times 3)}$ (Fig. \ref{fig:fp-representation}, Left). 

\paragraph{Semantic image} Floor plans are \textit{structured arrangements} consisting of subdivisions -- areas such as a living room or bathroom, openings such as doors or windows, and separators such as walls or railings. All floor plans that have categorical labels for subdivision areas and/or semantic information encoded at pixel level are referred to as \textit{semantic images} in our definition, $X \in \mathcal{X} = \{0, 1, \ldots, N_{c}\}^{H \times W}$, in which $N_{c}$ is the number of subdivision categories (Fig. \ref{fig:fp-representation}, Center). In the semantic image representation, pixel values correspond to subdivision categories \eg 1 $\leftarrow$ "living room", 2 $\leftarrow$ "bathroom", 16 $\leftarrow$ "internal wall", etc.

\paragraph{Access graph} Floor plans often have clearly identifiable \textit{relations} between subdivisions \eg access information between two rooms by means of a door. Even though semantic images efficiently capture the shapes and categories of the subdivisions, semantic image maps do not explicitly define the relations between the subdivisions. 

Instead, floor plans can be attributed an \textit{access graph}, directly laying bare the topological structure of the subdivisions (Fig. \ref{fig:fp-representation}, Right). Formally, a graph is defined as a 4-tuple $g = \left( N, E, \mu, \nu \right)$, in which $N$ is a finite set of nodes, $E$ is the set of edges, and $\mu$ and $\nu$ are the node and edge labelling functions respectively \cite{riba_learning_2021}. The nodes in an access graph correspond to the areas, and the edges to connections between the areas. 

Depending on the use-case and feature accessibility, the nodes and edges can have (one or several) attributes. Usually, room-type information is encoded on the nodes, and edges are only present if a door is in between two rooms, hence edges do not have an attribute \cite{nauata_house-gan_2020, shabani_housediffusion_2022, tang_graph_2023}. Even though we agree that access connectivity is the most important relation between rooms, an important relation is disregarded: room adjacency. Room adjacency is an influential factor \eg for decisions around privacy, structural integrity, and function. Access graphs, in the remainder of this work, have therefor a connectivity-type attribute, which is either "door" or "adjacent".

Proposed by \cite{vedaldi_learning_2020}, it is noteworthy to mention that several approaches \textit{compute} the edge features. Specifically, edge features between two nodes are computed by pairwise geometric features, such as the ratio between areas or the relative position.

%% INTERSECTION-OVER-UNION
\subsection{Intersection-over-Union}\label{sec:3_iou}

\begin{figure*}
\begin{center}
\includegraphics[width=0.95\textwidth]{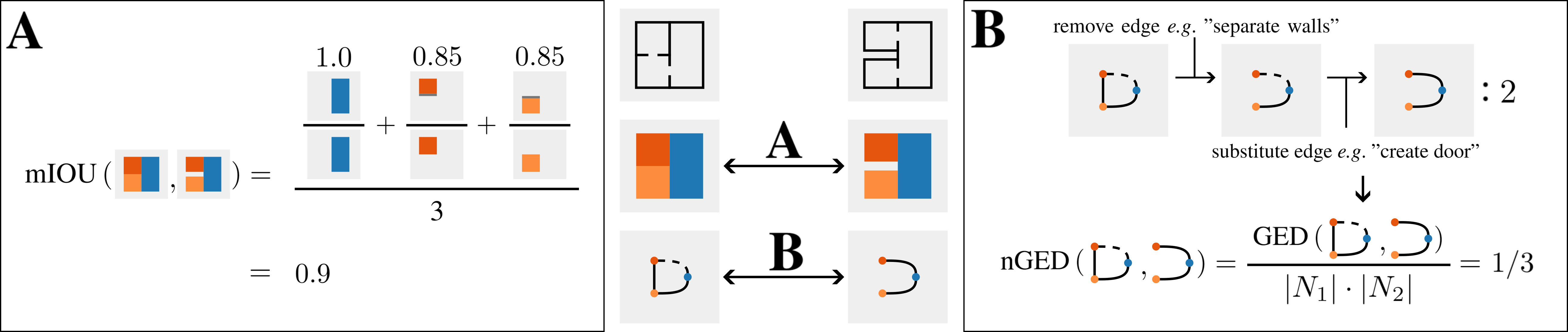}
\end{center}
   \caption{
   \textbf{Visual- and graph-based similarity metrics for floor plans}. A) IoU can be directly used as a measure for similarity between semantic image representations. mIoU between the semantic images is computed through Eq. \ref{eq:miou} (floor plans have 3 semantic labels); in this case being 0.90 which is a relatively high number. B) nGED is computed through Eqs. \ref{eq:ged-simple} and \ref{eq:nged}: GED equals the shortest edit path which is 2; the orders of the graphs are both 3, hence nGED $=2/9$. When $\gamma = 0.4$ (a typical value), $1 - \text{nGED} \left( g_1, g_2 \right)^\gamma \approx 0.45$. The visual-based similarity ($0.90$) and graph-based ($0.45$) similarity scores are significantly different which indicates that in some cases IoU and GED negatively correlate in "seeing" similarity.}
\label{fig:ssig-method}
\end{figure*}

The \textit{Intersection-over-Union} (IoU) is not considered a proper distance metric, but it is a well-adopted evaluation metric for tasks where measuring the overlap between regions or sets is important, such as in object detection or segmentation tasks.
In turn, IoU is often used as a measure for estimating structural similarity in various applications related to floor plans \eg for finding similar or matching floor plan pairs for similarity learning \cite{patil_layoutgmn_2021, jin_shrag_2022},  evaluating the accuracy of segmentation algorithms \cite{liu_raster--vector_2017}, or to check the diversity of outputs in floor plan generation \cite{nauata_house-gan_2020}. The IoU is commonly defined as the fraction between the amount of overlap and the union of two binary images $X_1$ and $X_2$ $\in \{0, 1\}^{H \times W}$: 

\begin{equation}\label{eq:iou}
    \text{IoU} \left( X_1, X_2\right) = \frac{X_1 \cap X_2}{X_1 \cup X_2}.
\end{equation}

When more classes are present in an image \eg in the case of semantic images ($X_1, X_2 \in \mathcal{X}$), IoU is computed per class and the results are aggregated to find the final score. For example, the mean IoU (mIoU) computes the average IoU, with respect to the ground truth, across the different classes, therefore treating every class similarly regardless of its proportion to the image size.

% remove edge \eg "separate walls", \newline substitute edge \eg "create door"

\begin{equation}\label{eq:miou}
    \text{mIOU} \left( X_1, X_2\right) = \frac{1}{N_c} \sum_{c=1}^{N_c} \text{IoU} \left( X_1==c,  X_2==c\right),
\end{equation}

\noindent where $N_c$ is the number of classes.  

Even though IoU is a powerful metric that in part is suitable for approximating structural similarity between floor plans, IoU frequently fails on both extremes: 1) false-negative: low IoU - high structural similarity, and 2) false-positive: high IoU - low structural similarity. 1) IoU is over-sensitive to irrelevant geometric perturbations in the image \eg translation, rotations, and scale. For example, when identical floor plans are centered at different locations in an image, the IoU will be low; falsely indicating that the floor plans are \textit{not} structurally similar. 2) IoU frequently fails to correctly identify the connectivity in floor plans \eg small "air gaps" between rooms are hardly identified by IoU yet are indicative of structural characteristics (see example in Fig. \ref{fig:ssig-method}).   

%% GRAPH-EDIT DISTANCE
\subsection{Graph Edit Distance}\label{sec:3_ged}

The \textit{Graph Edit Distance} (GED) \cite{sanfeliu_distance_1983}, is another measure to judge similarity between floor plans, used for \eg compatibility of generated layouts \cite{nauata_house-gan_2021} or for floor plan retrieval \cite{park_floor_2023}. GED is a metric that quantifies the similarity (dissimilarity) between two graphs by computing the minimum cost of transforming one graph $g_1$ into (a graph isomorphic to) $g_2$ in terms of a sequence of edit operations $\left( e_1, e_2, \ldots, e_j \right)$, referred to as the \textit{edit path}. Edit operations typically include node and edge deletions, insertions, and substitutions, which can have different costs. Given a set of graph edit operations $e_i$ and associated costs $c_i \left(e_i\right)$, the $\text{GED} \left( g_1, g_2 \right)$ is formally defined as,

\begin{equation}\label{eq:ged}
    \text{GED} \left( g_1, g_2 \right) = \min_{\left( e_1, e_2, \ldots, e_j \right) \in \pi \left(g_1, g_2\right)} \sum_{j=1}^k c_j \left( e_j \right),
\end{equation}

\noindent in which $\pi \left(g_1, g_2\right)$ denotes the set of edit paths that transforms $g_1$ into (a graph isomorphic to) $g_2$. In our work, all edit operations have the same cost, hence Eq. \ref{eq:ged} becomes,

\begin{equation}\label{eq:ged-simple}
    \text{GED} \left( g_1, g_2 \right) = \min_{k} \left( e_1, e_2, \ldots, e_k \right),
\end{equation}

\noindent such that $\left( e_1, e_2, \ldots, e_k \right) \in \pi \left(g_1, g_2\right)$. GED in this case is equivalent to the smallest number of edit operations. To normalize the GED between 0 and 1, we define the \textit{Normalized Graph Edit Distance} (nGED) by dividing GED with the product of the orders\footnote{Note that Eq. \ref{eq:nged} does not strictly normalizes GED. Even so, for floor plans \eg, in \cite{wu_data-driven_2019}, nGED will always be below 1. Furthermore, the product of the orders, instead of for example the sum, nicely spreads the nGED distributions.} -- the number of nodes -- of the graphs ($\left| N_1 \right|$ and $\left| N_2 \right|$):

\begin{equation}\label{eq:nged}
    \text{nGED} \left( g_1, g_2 \right) = \frac{\text{GED} \left( g_1, g_2 \right)}{\left| N_1 \right| \cdot \left| N_2 \right|}.
\end{equation} 

\begin{figure}[h!]
    \centering
    \includegraphics[width=0.95\linewidth]{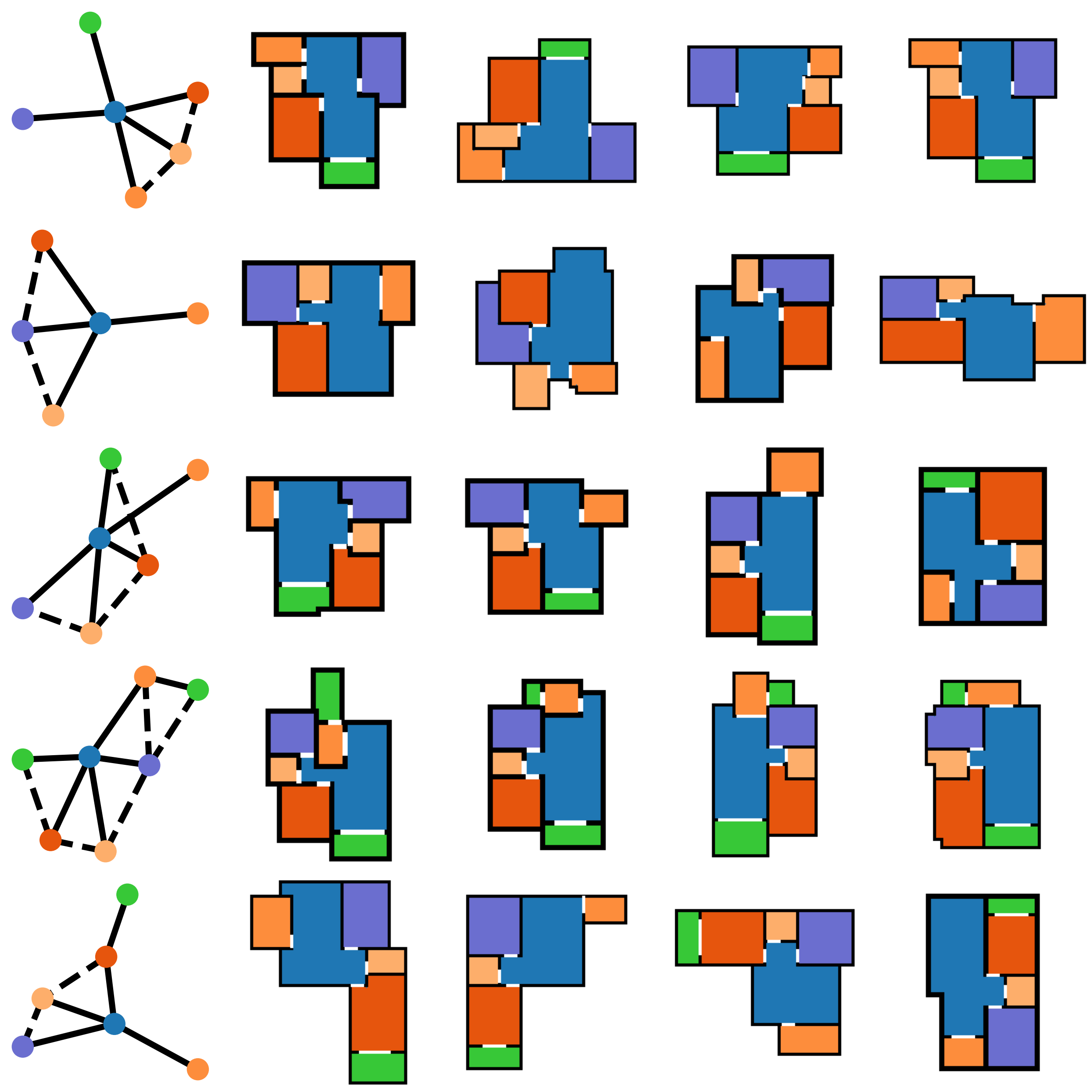}
    \caption{
    \textbf{Graph to image examples.} 
    Each row represents an access graph (column 1) and several randomly selected corresponding floor plans semantic images (columns 2 - 5).
    }
    \label{fig:ged}
\end{figure}

% TODO Based on Review
% \todo[inline]{[Review-based To Do] Why the multiplication of the orders of the graph?}
% L 402-403: “the product of the orders – the number of nodes – of the graphs (|N1| and |N2|)” : Can you provide some insights on why you choose this as the normalization denominator? This can play a significant role in the GED values. I would expect that e.g., a sum of nodes might be more appropriate, especially when floor plans (and their graphs) are very large, and some consideration on edges as well. Although I agree with the insights in L 584-640, I am unsure to which degree the specific formulation of the normalized GED plays a role. For example, an almost maximum of 0.5 for the entire dataset might point to that fact. A mathematical explanation or some strong insights on the choice of normalization would be important to add.

nGED can be used as of measure of closeness between the topology of two floor plans, in which a small nGED suggests that two floor plans share topological characteristics, which could be, for example, indicative of commonalities in flow or function between two buildings. 

nGED has several limitations. First of all, the amount of different floor plans possible for a particular access graph is huge, for which many corresponding floor plans pairs are not structurally similar (see Fig. \ref{fig:ged}). Second, it is unclear which edit costs align with structural similarity: Is the removal of a node or an edge more costly? Does the node label matter? etc. Third, nGED is compute-heavy which limits its use in real-time systems \eg floor plan retrieval in search engines. We elaborate more on the shortcomings in Subsec. \ref{subsec:distributions}.

% TODO Based on Review
% \todo[inline]{[Review-based To Do] Enrich graph representation with shape / visual information, why doesn't this work instead?}
% It seems the GED distance is not rich enough. The proposed score actually considers only GED and IOU in isolation. Is it not possible to enrich the graph representation with information from the IOU? For example, can we not add the size (in m^2) to the node's features? Etc.? Is this a more principled approach then combining IOU and GED separately? Or is this not computationally feasible, or is it problematic to define the cost for the GED in that case?

In Subsec. \ref{subsec:correlations} we investigate the trends and correlations between nGED and mIoU on RPLAN \cite{wu_data-driven_2019} empirically.

%% A VISUALLY-GUIDED GRAPH-EDIT DISTANCE
\subsection{A Visually-guided Graph Edit Distance}\label{sec:3_ssig}

Independently, IoU and GED fail to holistically capture structural similarity between floor plans. Therefore, instead of treating images and graphs separately, we propose a metric that ascertains closeness based on the semantic image \textit{and} access graph. We define the similarity between a pair of floor plans $(p_1, p_2)$, $p_i = (X_i, g_i) \in \mathcal{X} \times \mathcal{G}$, as

\begin{equation}
    \label{eq:ssig}
    \text{SSIG} \left( p_1, p_2 \right) = \frac{\text{mIoU} \left( X_1, X_2 \right) + \left( 1 - \text{nGED} \left( g_1, g_2 \right)^\gamma\right)}{2}.
\end{equation}

Coined as \textbf{SSIG} (\textbf{S}tructural \textbf{S}imilarity by \textbf{I}oU and \textbf{G}ED), \ref{eq:ssig} essentially computes a weighted average between a IoU-based and a nGED-based similarity metric. $\gamma \in \mathbb{R}^+$ is a weight that allows to tune between the relevance of nGED w.r.t. mIoU: the higher $\gamma$ the \textit{less} influence nGED and vice versa.\footnote{Note that nGED $\in (0, 1)$, hence increasing the power term decreases nGED.} For example for RPLAN \cite{wu_data-driven_2019}, the mIoU and nGED distributions are balanced when $\gamma = 0.4$. Specifically, balanced distributions here means that the IoU and nGED probability density functions have as much overlap as possible.

% TODO Based on Review
% \todo[inline]{[Review-based To Do] Add how to tune gamma and why needed.}
% The gamma value could potentially mitigate some of the above. However, you do not discuss how you set the value of gamma although you mention in L 465 that you would describe this in Section 4 (“the IoU and GED distributions are balanced when γ = 0.4 (see Sec. 4 for more details)”).

% + 

% It is not clear how the trade-off parameter gamma was tuned, calling into question whether it was tuned based on results obtained by eye?

%% ALGORITHM AND EXPERIMENTAL SETUP
\subsection{Efficient algorithm for ranking}\label{sec:3_algo}

Even for small graphs, GED is a compute-heavy measure, usually taking several seconds per floor plan pair. When \textbf{SSIG} is used for evaluation, pair mining, or direct floor plan retrieval on large datasets, the time constrain hugely limits practical usage. We propose an algorithm that drastically speeds up retrieval and finds structurally similar pairs relatively fast. 

First, mIoU is computed for every combination of floor plan pairs in the dataset. Second, only the $n$ (set between 50 and 100) best scoring pairs on mIoU for each floor plan identity are kept. Third, the \textbf{SSIG} is computed for each remaining pair. Fourth, the corresponding lists for each floor plan identity are ranked on \textbf{SSIG} in descending order. 

The algorithm assumes that the best matches (highest \textbf{SSIG} scores) are found in the $n$-best mIoU pairs. In the next session, we empirically find that the assumption generally holds if $n$ is large enough. In Subsec. \ref{subsec:retrieval}, we show that for ranking RPLAN, the assumption holds for relatively small $n$. For a dataset with size $N$ (usually between 10k and 100k), the algorithm is orders of magnitude faster than brute-forcing through it.

An assessment of the ranking algorithm and relation to floor plan retrieval systems are provided in the following section (Sec. \ref{sec:results}).

\section{Results and Evaluation}\label{sec:results}

\subsection{Data}
In this paper, RPLAN \cite{wu_data-driven_2019} is used for analysis and evaluation. RPLAN is a large-scale dataset containing floor plan images with semantically segmented areas, consisting of $>$80k single-unit apartments across Asia. We further cleaned the dataset, where there were several apartments with rooms without any doors \ie disconnected in terms of the topology. Moreover, apartments are removed that are not fully connected and/or for which our graph extraction algorithm cannot reliably compute the corresponding access graphs. The cleaned dataset contains $\sim$56k apartments. 
Furthermore, we found that RPLAN contains many (near) duplicates. For evaluation of the ranking algorithm (Subsec. \ref{subsec:retrieval}), we remove the (near) duplicates to reduce clutter in the results. Duplicates are removed by a threshold on mIoU, $\tau_{\text{IoU}} = 0.87$. Duplicates are \textit{not} removed when assessing the distributions and correlations of mIoU and nGED. 

\subsection{Distributions}\label{subsec:distributions}

To understand the relations between the pairwise similarity measures (mIoU and nGED) and how they relate to structural similarity, we start by investigating the probability density functions of the pairwise similarity measures ($p_{\text{IoU}}$ and $p_{\text{GED}}$, respectively). $p_{\text{IoU}}$ and $p_{\text{GED}}$ are approximated by computing IoU and GED for over a million randomly sampled floor plan pairs. For brevity, we use "IoU" and "GED" instead of "mIoU" and "nGED".

\begin{figure}[h!]
    \centering
    \includegraphics[width=0.90\linewidth]{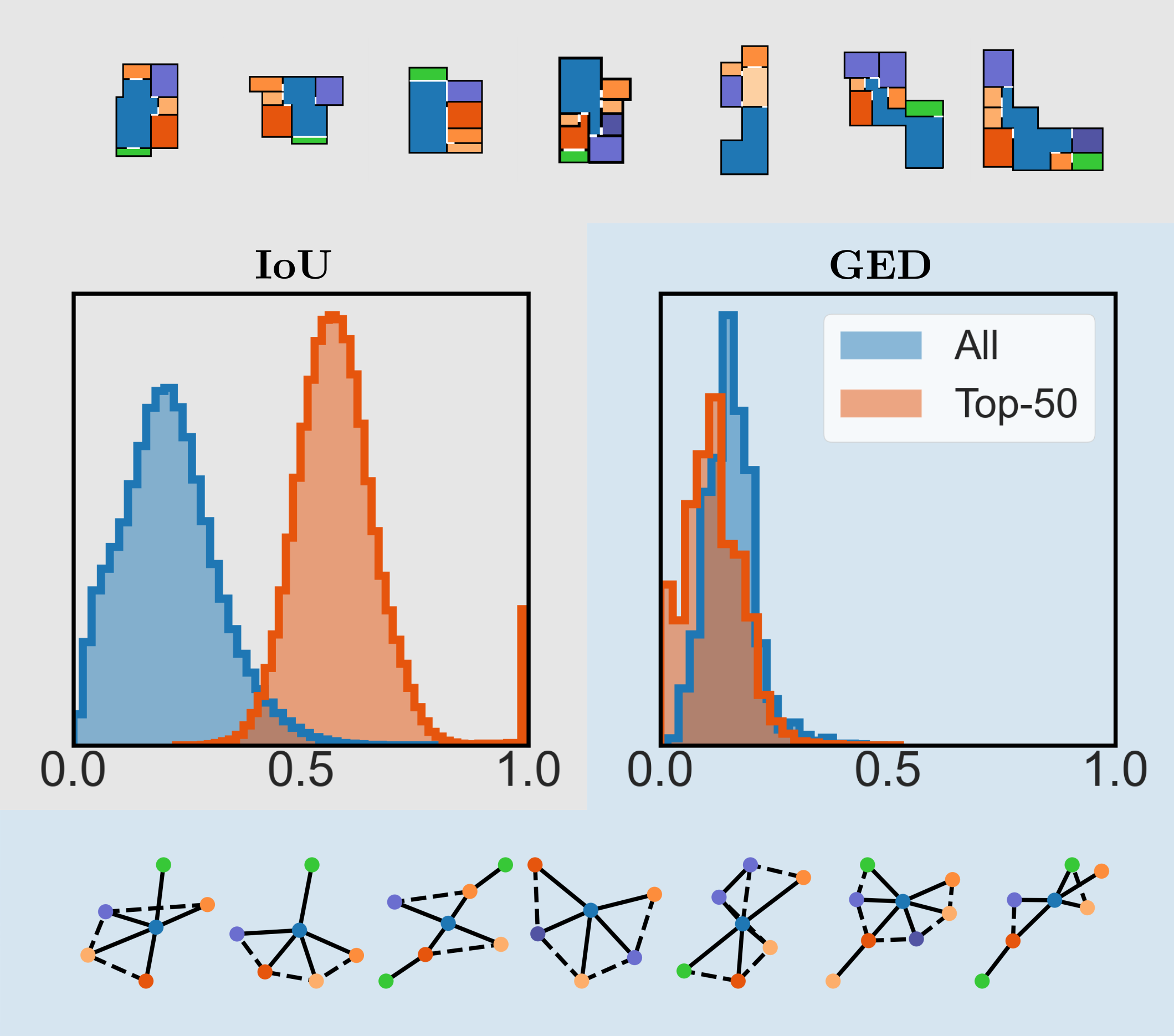}
    \caption{
    \textbf{IoU and GED distributions and originality.} The two plots in the center illustrate the distributions of IoU (left) and GED (right): blue depicts the distribution over all samples and orange only considers the top-50 \textit{on IoU}. An indication of semantic image originality (based on the average IoU) is given on the top and one for the graph originality (based on the base graph occurrence) is given on the bottom: originality increases going from left to right.
    }
    \label{fig:distributions}
\end{figure}

\paragraph{IoU} IoU is computed according to Eq. \ref{eq:miou} and across all subdivision categories except for the "background". The histogram of the distribution is provided in Fig. \ref{fig:distributions} in the middle left plot in blue. The mean of the distribution is around 0.25 which means that on average approximately 25\% of the pixels are overlapping. We also plot the IoU distribution for the 50 best-scoring examples on IoU per floor plan identity, shown in orange. As expected, compared to the full IoU distribution, the mean is significantly shifted. The spike around 1.0 can be attributed to the many (near) duplicates in the dataset. 

To give an idea of the diversity of floor plan samples, Fig. \ref{fig:distributions} (top row) highlights floor plans with different 'originality' scores. The originality is computed by the average IoU score for a given sample with all remaining samples. The more right ($\approx$ 0.1) the more original \ie seldom and the more left ($\approx$ 0.4) the less original \ie more frequent. Generally speaking, the originality positively correlated mostly with the complexity of the shapes of both the rooms and boundaries.

% TODO Based on Review
% \todo[inline]{[Review-based To Do] Add IoU Originality Threshold}
% L 542: “The originality is computed by the average IoU score” —> average over all found pairs? What it the threshold of IoU to consider they are pairs?

\paragraph{GED} nGED is computed according to Eqs. \ref{eq:ged-simple} and \ref{eq:nged} and hence cost for node and edge deletions, insertions, and substitutions are kept the same. The histogram of the full distribution is illustrated in Fig. \ref{fig:distributions} in the middle right plot in blue. We also plot the GED distribution for the 50 best scoring examples \textit{on IoU} per floor plan identity, shown in orange. Surprisingly, both full and top-50 distributions are nearly similar, revealing a weak (if at all) correlation between IoU and GED. The weak correlation is further investigated in the next section.

\begin{figure}[h!]
    \centering
    \includegraphics[width=0.90\linewidth]{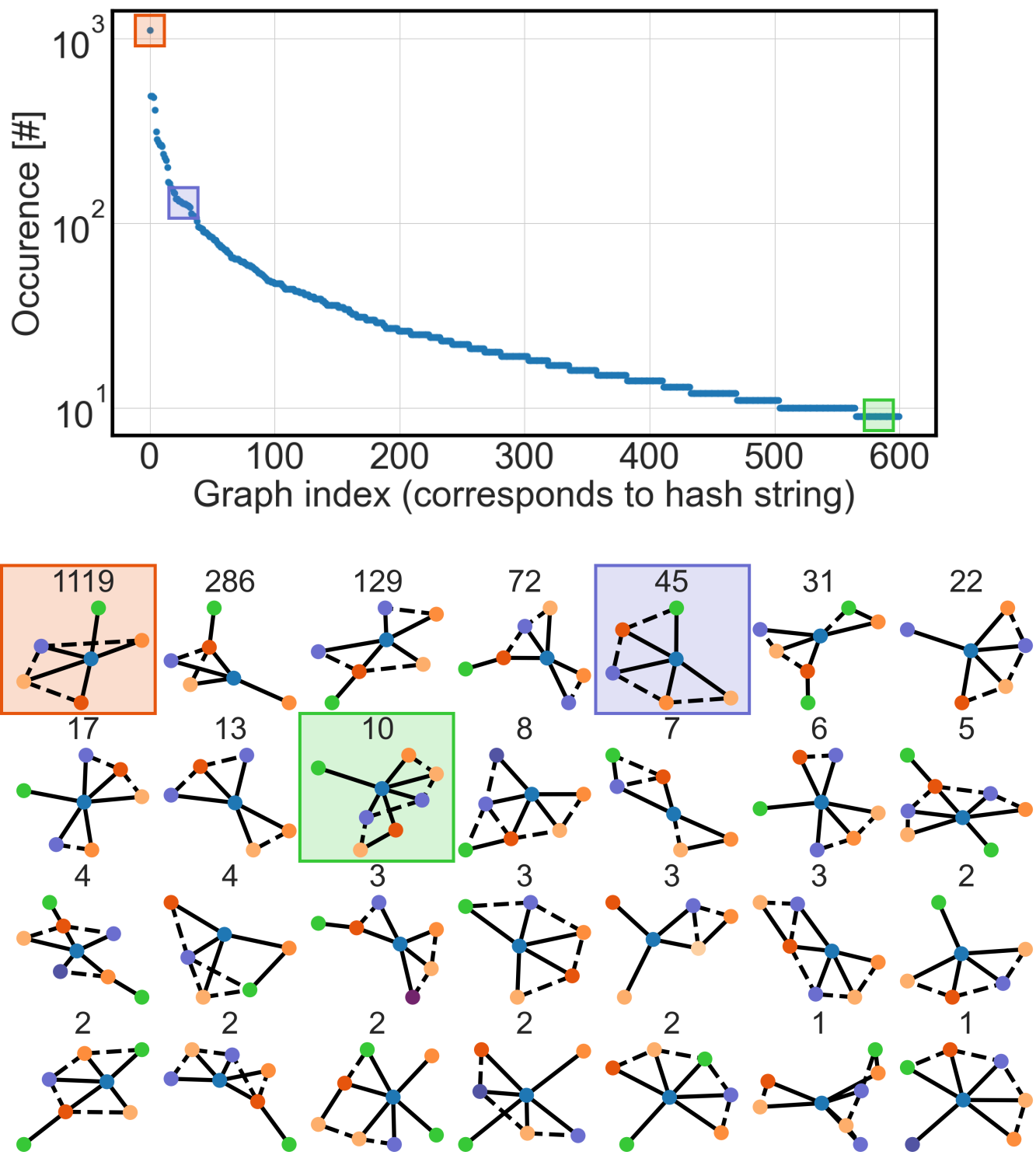}
    \caption{
    \textbf{Distribution of base graphs.} The plot on the top provides the distribution of base graphs (in descending order of occurrence) for base graphs that at least occur 9 times. On the bottom, the base graphs are plotted for several randomly picked floor plans. For some of the base graphs, we indicate the position in the distribution through the colored boxes.
    }
    \label{fig:base-graphs}
\end{figure}

It is worth noting that there are only 50 unique GED values in our population of random pairs. The limited range of GED values can be attributed to: 

Firstly, the amount of floor plan pairs that share the same underlying \textit{base graph} -- all possible graphs present in the dataset for which each pairwise combination is isomorphic: $g_1$ is isomorphic to $g_2$ -- is relatively large. To make our point, we sort all base graphs based on the number of corresponding floor plans and show the huge imbalance between base graph occurrences (see Fig. \ref{fig:base-graphs}). For example, more than 25\% of the data ($\approx $14\text{k} samples) correspond to the 100 most prominent base graphs only. Visualized in Fig. \ref{fig:distributions} (bottom row) and similar to the IoU originality scores, we score floor plans on graph originality as well. Graph originality is based on the number of corresponding floor plans for a given base graph: the more original the fewer floor plans per base graph. While the main patterns of graph originality are to be more closely investigated, on average the number of nodes as well as edges slightly increase with originality.

% ---
% TODO Based on Review
% \todo[inline]{[Review-based To Do] Add Graph Originality explanation.}
% Please provide an insight on graph originality as you did for IoU (eg the more to the right, the …more nodes? More edges? )
% ---

Secondly, revealed empirically by the graph originality distribution, the distribution of topology -- the graph's structure -- alone is small because the graphs are limited in size (number of connections) and order (number of subdivisions). Furthermore, the topological \textit{types} are limited as well. Typically, access graphs are confined to be flower-like and thus centered around a single node. The central node is usually the living room and/or hallway.

Indeed, the majority of floor plans stem from a limited collection of base graphs, and only a few node and edge attributes are considered, leading to the ineffectiveness of measuring structural similarity solely based on graph similarity.

\subsection{Correlations}\label{subsec:correlations}

\begin{figure}[h!]
    \centering
    \includegraphics[width=0.90\linewidth]{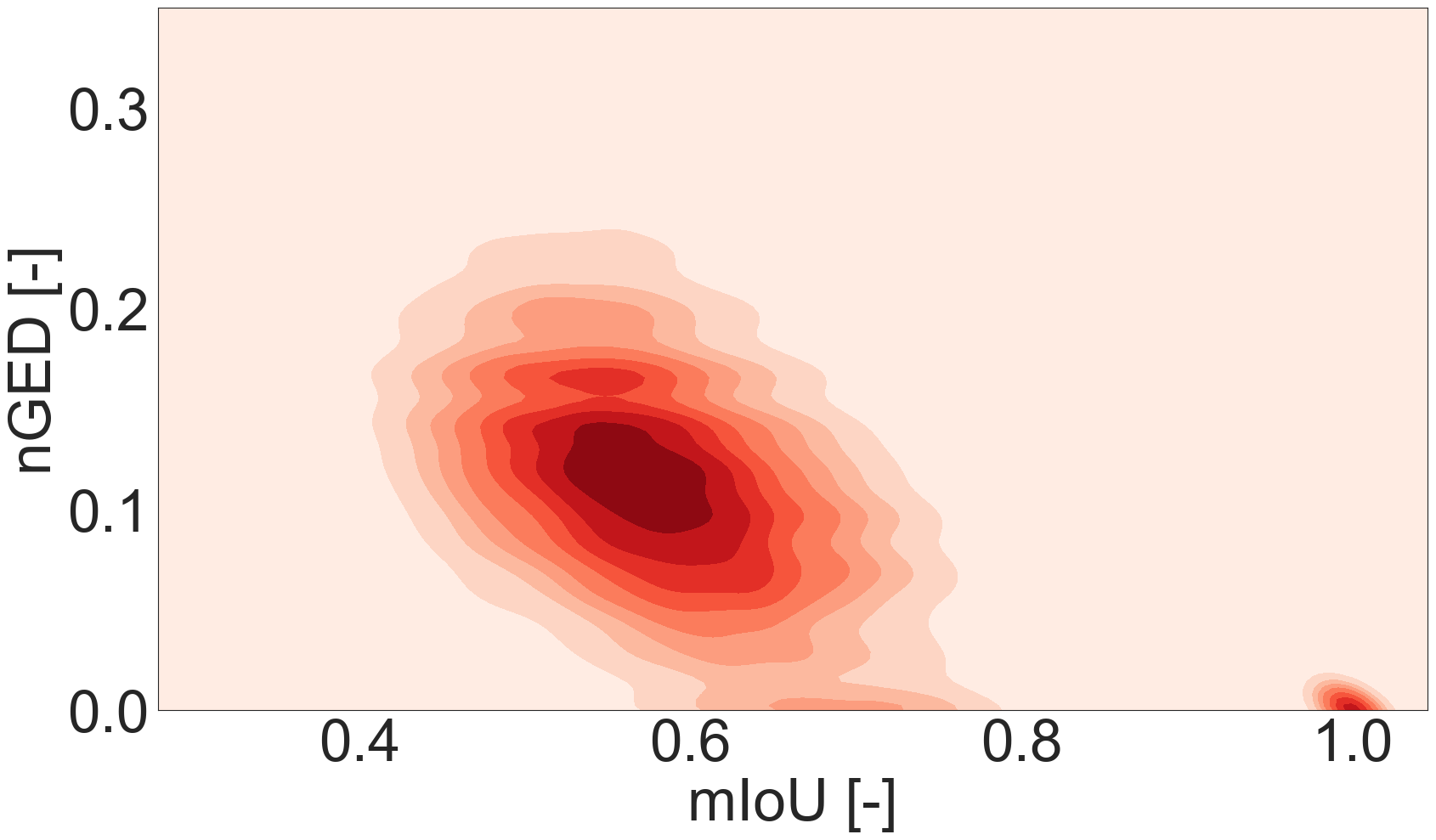}
    \caption{
    \textbf{2D density map of IoU and GED as heatmap.} A moderate correlation (0.503) between IoU and -GED is found. The distribution is, however, widely spread indicating that IoU and GED often fail to predict one another with confidence.}
    \label{fig:correlations}
\end{figure}

\begin{figure}[h!]
    \centering
    \includegraphics[width=0.70\linewidth]{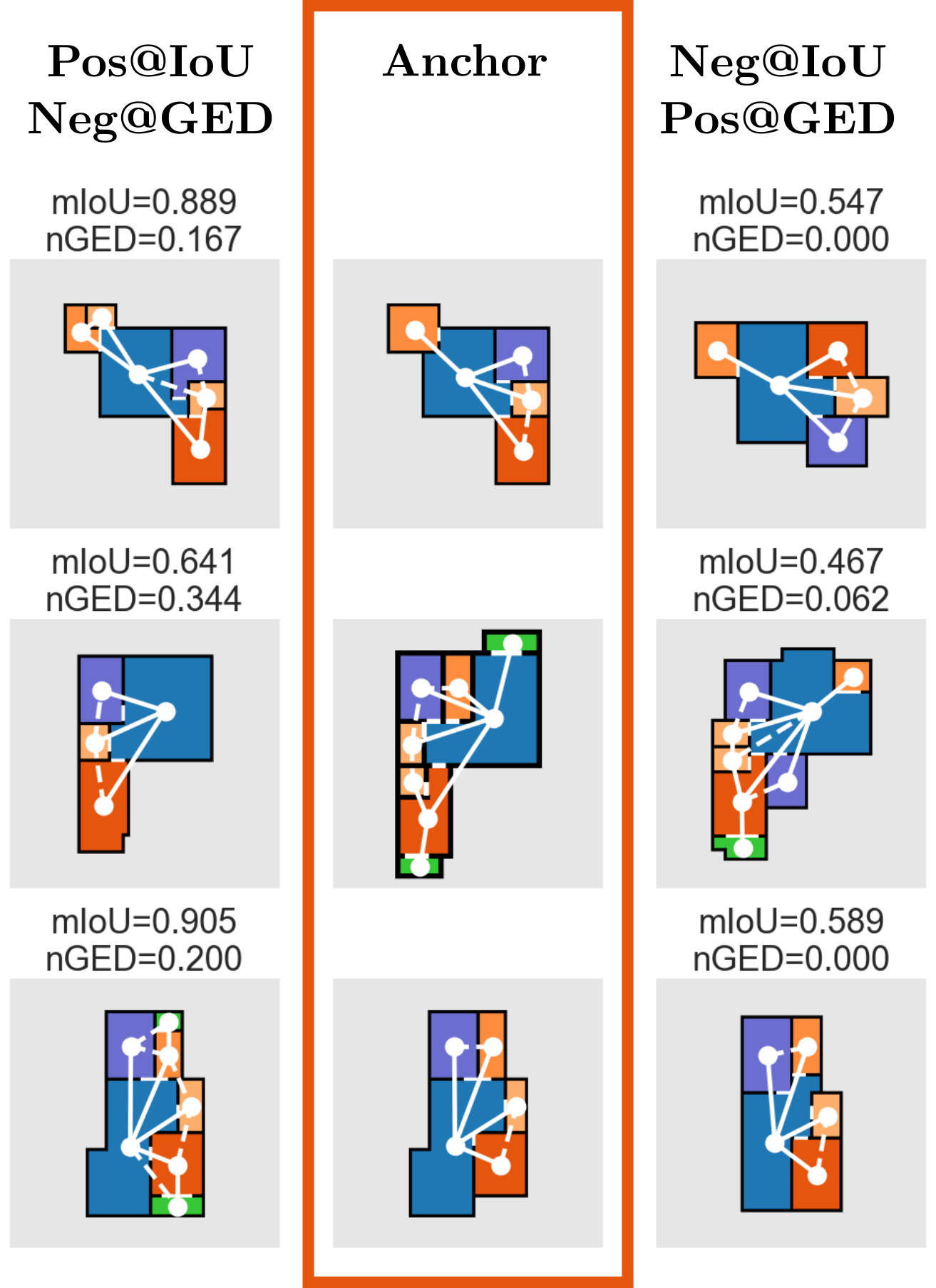}
    \caption{
    \textbf{Negative examples between IoU and GED on RPLAN.} IoU is sensitive to the scales and shape of the room while GED captures the connections between the rooms. 
    }
    \label{fig:extremes}
\end{figure}

The correlation between IoU and GED is further investigated. The 2D density map of IoU and GED is depicted in Fig. \ref{fig:correlations} and plotted as a heatmap. We define a positive correlation when increments in visual similarity \ie \textit{higher} IoU result in increments in the graph-based similarity \ie \textit{lower} GED. Effectively we measure the correlation between IoU and \textit{negative} GED. The heatmap shows for increasing IoU ranges that GED distributions shift towards lower values, hence the correlation between IoU and GED is positive. To be precise, the Pearson correlation coefficient is 0.503 between mIoU and -nGED, hence GED and IoU are moderately correlated. However, the overlapping regions between GED distributions for different IoU ranges are often large. Specifically, IoU and GED oppose each other 38\% of the time -- opposition here means that a higher IoU results in a higher GED or that a lower IoU results in a lower GED. Several examples in which IoU and GED oppose each other are given in Fig. \ref{fig:extremes}.

From the triplets shown in Fig. \ref{fig:extremes}, many characteristics of IoU and GED rise to the surface. For an anchor, we define pos@IoU as the positive (IoU high) and neg@IoU as the negative query (IoU low). The same holds for pos@GED (GED low) and neg@GED (GED high). Most prominent is the sensitivity of IoU w.r.t. the overall shape of the building: in all cases, the overall shape between the anchor and the pos@IoU are nearly identical and slightly different for the pos@GED. On the other hand, the number of rooms is often different between the anchor and pos@IoU while (nearly) always equivalent between the anchor and pos@GED: placing or removing rooms is detrimental for GED but in many cases does not hurt IoU so much. Another difference between pos@IoU and pos@GED is the fact that changes in the room shapes -- especially to the central space -- are \textit{not} penalized by GED yet detrimental for IoU. The examples reveal the difficulty of measuring the structural similarity of floor plans. Apparently, both measures independently cannot fully grasp structural similarity.

% ---
% TODO Based on Review
% \todo[inline]{[Review-based To Do] SSIG limitations.}
% L 737-739: “is a proper first step to find correspondences in structural similarity coarsely, while GED is a proper help to measure structural similarity on a finer level.” —> However, in L 684 you mention “the sensitivity of IoU w.r.t. the overall shape of the building”. This points that if using IoU for a first search, the selected floorpans will be very similar on the shape of the rooms. I am wondering if that limits the outputs especially when setting as a premise “floor plan similarity”. I do understand the issue of computation and this is a good compromise. I would suggest reminding the reader of this, but also making a direct connection to Figure 8 that show the potential of ending up with dissimilar shapes as well.
% ---

% SUBSECTION: FLOOR PLAN RETRIEVAL
\subsection{Floor plan retrieval}\label{subsec:retrieval}

% Structure of subsection
We evaluate our ranking algorithm provided in Subsec. \ref{sec:3_algo} and qualitatively compare the retrievals to independent IoU- and GED-based rankings, and to retrievals based on a deeply learned metric \cite{patil_layoutgmn_2021}.

\begin{figure}[h!]
    \centering
    \includegraphics[width=0.95\linewidth]{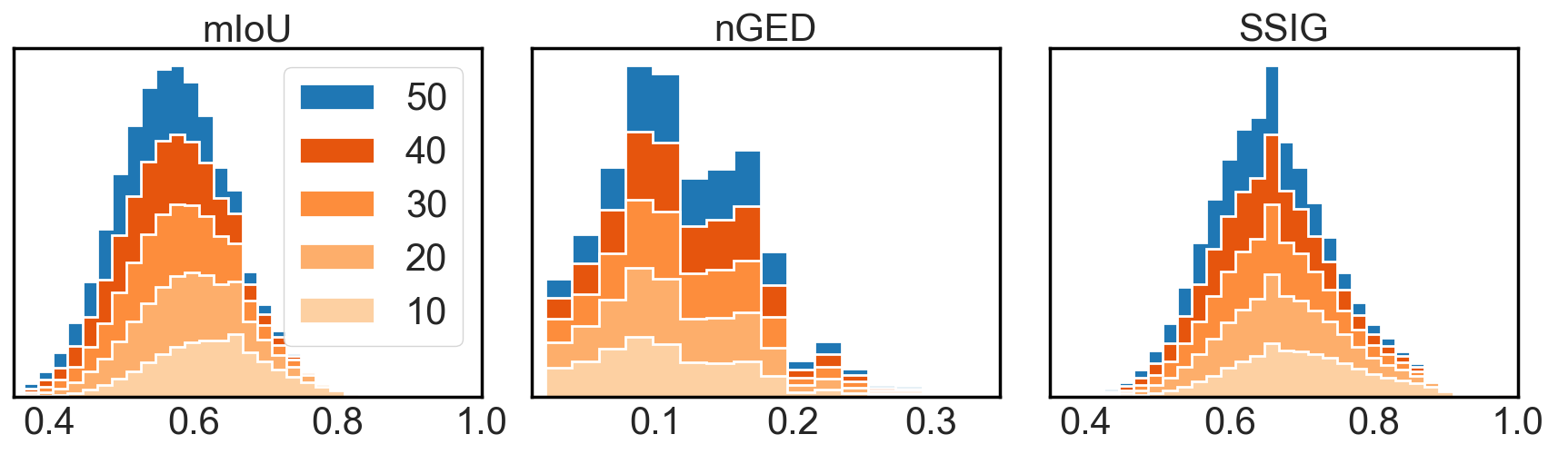}
    \caption{
    \textbf{IoU, GED, and SSIG distributions for various $n$}. The different colors indicate different $n$. Particularly interesting is the fact that the mean of both IoU and SSIG drastically decreases for increasing $n$. 
    }
    \label{fig:distributions-top}
\end{figure}

% Settings of the algorithm
We start by finding a proper value for $n$ -- the number of samples (when ranked on IoU) considered for further \textbf{SSIG} assessment. In Fig. \ref{fig:distributions-top}, the distributions for IoU, GED, and \textbf{SSIG} are provided for various values of $n$. In our observation, for relatively large values of \textbf{SSIG} ($>$0.7) every linear increase in $n$ adds fewer samples every time. We found for RPLAN that values of $n$ $>$= 50 seldom change the top-10 on \textbf{SSIG}. Compared to measuring \textbf{SSIG} on the whole dataset, \textbf{SSIG} only needs to be computed 50 times. We believe that IoU, at least in the case of RPLAN, is a proper first step to find correspondences in structural similarity coarsely, while GED is a proper help to measure structural similarity on a finer level. Hence, we refer to our ranking method as a visually-guided graph edit distance.

% Figure retrieval results
\begin{figure*}[h!]
    \centering
    \includegraphics[width=0.95\textwidth]{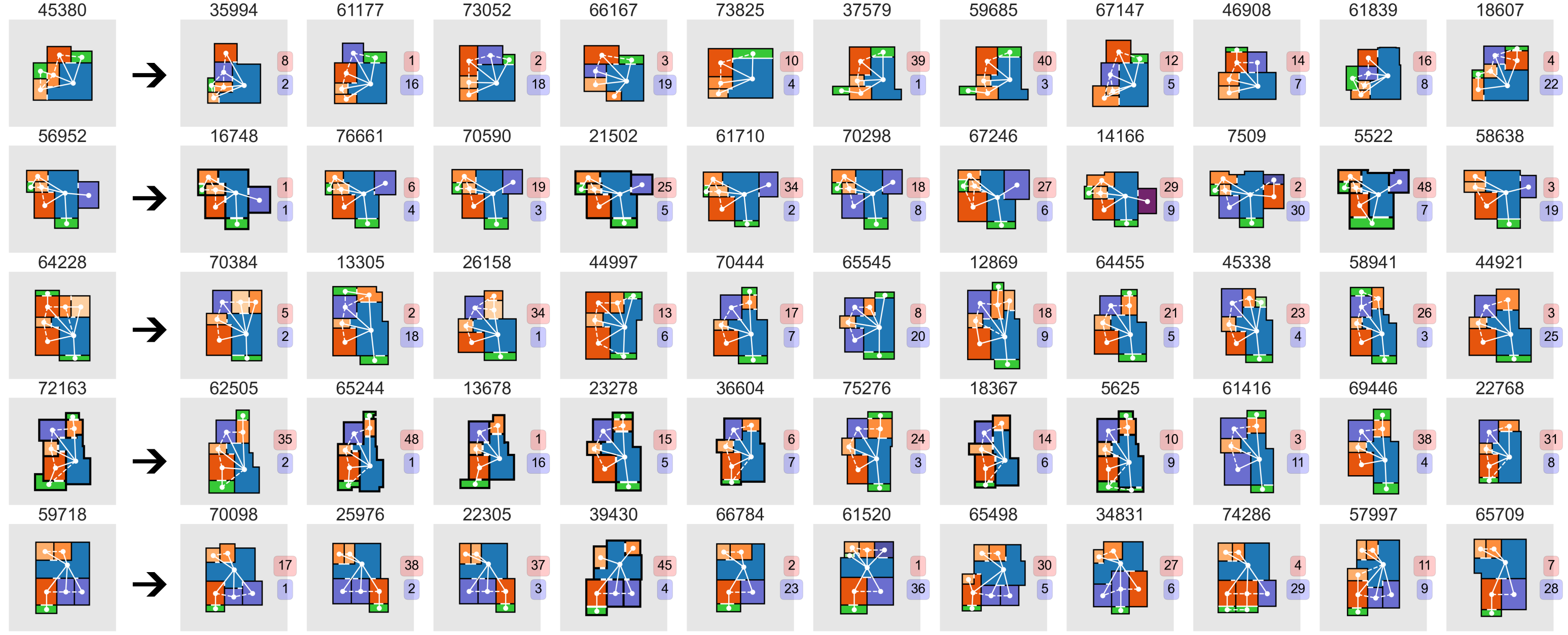}
    \caption{
    \textbf{Retrieval results of our method.} The most left is the query and to the right are the retrievals. The red and blue transparent boxes indicate the IoU and GED ranks, respectively.
    }
    \label{fig:ranks}
\end{figure*}

% Figure RPLAN similarity distribution for different n

% Ranking efficiency
Our algorithm is compute-efficient w.r.t. GMNs. Depending on the size and originality of the floor plan's corresponding access graphs, our algorithm takes between 1 and 5 s on a single CPU per unseen sample to fully rank it. Compared to LayoutGMN, the same computation takes approximately 1500 s. Hence, our algorithm is compute-efficient w.r.t. GMNs.

% Retrieval results
Randomly picked retrievals (top 10) of our method are shown in Fig. \ref{fig:ranks} and compared against IoU- and GED-only ranks. In red and blue are IoU and GED ranks, respectively. Not surprisingly the \textbf{SSIG}, IoU, and GED ranks are quite different which can be attributed to the fact that the correlation between IoU and GED is only weakly positive (see Subsec. \ref{subsec:correlations}). Compared to IoU- and GED-only retrievals, it can be seen that \textbf{SSIG} indeed compensates for 'failures' in IoU and GED. For example in the last row of Fig. \ref{fig:ranks}, the first 4 retrieval results, even though having a low IoU value, are compensated by GED, while the 5th retrieval is compensated by IoU, even though having a low GED value. Similar compensation for failures can be found back in the retrievals of deeply learned metrics, see \cite{patil_layoutgmn_2021, jin_shrag_2022, vedaldi_learning_2020}.

% ---
% TODO Based on Review
% \todo[inline]{[Review-based To Do] Explain lack of ground truth.}
% There is a lack of ground truth in the study, which ideally would be resolved via a user study
% ---

% ---
% TODO Based on Review
% \todo[inline]{[Review-based To Do] Explain the lack of quantitative comparisons.}
% The comparison with the state-of-the-art method is qualitative and not very detailed (only Fig. 8). Ideally, a quantative comparison would also be included (for example, how much do the rankings agree, etc.?). 

% +

% L 726-727: “IoU- and GED-based rankings, and to retrievals based on a deeply learned metric [13]” —> I do not think I saw any results on the deep learning metric from 13.
% ---
\section{Conclusion and future work}
% Successful data-driven methods that compute structural similarity between floor plans require cross-graph information sharing during training and inference. Feature vectors can, therefore, \textit{never} be computed offline, seriously hindering the practical usability of such approaches in various cases. We believe that the cross-graph information sharing is needed in such approaches because the framework's objective -- mimicking a IoU similarity score -- is not aligned with structural similarity in the first place. 

In this work, we showed that image- as well as graph-based similarity metrics alone are likely not enough to robustly measure structural similarity between floor plans. We overcome the shortcomings by combining IoU and GED into one metric, called \textbf{SSIG}. We qualitatively showed that a floor plan database which is ranked on \textbf{SSIG} shares similar characteristics to results of state-of-the-art methods that use GMNs to compute the ranking. 

This work is however limited to investigating IoU and GED and does not consider other well-known image- and graph-based similarity metrics. Moreover, we did not conduct any user study to rigorously assess our claims further. In future studies we hope to address both limitations. 

The formulation of similarity as defined in Eq. \ref{eq:sim} is setup in a task-agnostic manner - the formulation addresses floor plan similarity as a \textit{single} overall measure. In many cases, however, a more realistic way to setup similarity might be dynamic (\eg, based on categories), hence, allowing for different similarity scores across multiple use cases whether used in floor plan tailored search engines, evaluation of floor plan generation, etc. Our aim is to work on such formulations in the future.

We believe that \textbf{SSIG} will pave a way towards improved data-driven research on floor plans, such as for providing a more natural objective to train deep metric learning frameworks that do not necessarily require information sharing across branches. 

% ---
% TODO Based on Review
% \todo[inline]{[Review-based To Do] Similarity per task.}
% For a user study, it would be quite important to consider what is the task. It should be evident that a similarity measure for the retrieval can be very task-dependent. For example, if a construction company knows the number of houses and m^2, there may be clear constraints on the size of the generated layouts. Are there other cases where we can clearly identify desiderata for the similarity measure? Would it make sense to design a similarity measure per task / goal, or do you think it is possible to design an overall similarity measure that would work well for all tasks?
% ---

{\small
\bibliographystyle{ieee_fullname}
\bibliography{bibliography}
}

\end{document}